\documentclass{article}

\usepackage[main,final]{neurips_2026}

\usepackage[utf8]{inputenc} 
\usepackage[T1]{fontenc}    
\usepackage{hyperref}       
\usepackage{url}            
\usepackage{booktabs}       
\usepackage{amsfonts}       
\usepackage{nicefrac}       
\usepackage{microtype}      
\usepackage{xcolor}         

\usepackage{amsmath}
\usepackage{amssymb}
\usepackage{mathtools}
\usepackage{amsthm}
\usepackage{wrapfig}
\usepackage{tabularx}
\newcolumntype{Y}{>{\raggedright\arraybackslash}X}

\usepackage[capitalize,noabbrev]{cleveref}

\theoremstyle{plain}
\newtheorem{theorem}{Theorem}[section]
\newtheorem{proposition}[theorem]{Proposition}
\newtheorem{lemma}[theorem]{Lemma}
\newtheorem{corollary}[theorem]{Corollary}
\theoremstyle{definition}

\theoremstyle{remark}

\usepackage{algorithm}
\usepackage{algorithmicx}
\usepackage{algpseudocode}

\usepackage{microtype}
\usepackage{graphicx}
\usepackage{subcaption}
\usepackage{booktabs} 
\usepackage{multirow} 
\usepackage{makecell} 
\usepackage[table]{xcolor}
\definecolor{lightblue}{RGB}{235,242,252}
\usepackage{hyperref}

\title{World–Value–Action Model: Implicit Planning for Vision–Language–Action Systems
}

\author{
Runze Li\thanks{Equal contribution.} \\
Westlake University \\
Hangzhou, China \\
\And
Hongyin Zhang\footnotemark[1] \\
Westlake University \\
Hangzhou, China \\
\And
Junxi Jin \\
Westlake University \\
Hangzhou, China \\
\AND
Qixin Zeng \\
Westlake University \\
Hangzhou, China \\
\And
Zifeng Zhuang \\
Westlake University \\
Hangzhou, China \\
\And
Yiqi Tang \\
Westlake University \\
Hangzhou, China \\
\AND
Shangke Lyu\textsuperscript{$\dagger$} \\
Nanjing University \\
Suzhou, China \\
\And
Donglin Wang\textsuperscript{$\dagger$} \\
Westlake University \\
Hangzhou, China \\
}

\begin{document}

\maketitle

\begingroup
\renewcommand{\thefootnote}{}
\footnotetext{%
\textsuperscript{$\dagger$}Corresponding authors:
\texttt{shangke\_lyu@nju.edu.cn}
\qquad
\texttt{wangdonglin@westlake.edu.cn}
}
\endgroup


\begin{abstract}
Vision--Language--Action (VLA) models have emerged as a promising paradigm for building embodied agents that ground perception and language into action. 
However, most existing approaches rely on direct action prediction, lacking the ability to reason over long-horizon trajectories and evaluate their consequences, which limits performance in complex decision-making tasks.
In this work, we introduce \textbf{World--Value--Action (WAV)} model, a unified framework that enables implicit planning in VLA systems. 
Rather than performing explicit trajectory optimization, WAV model learn a structured latent representation of future trajectories conditioned on visual observations and language instructions. 
A learned world model predicts future states, while a trajectory value function evaluates their long-horizon utility. 
Action generation is then formulated as inference in this latent space, where the model progressively concentrates probability mass on high-value and dynamically feasible trajectories.
We provide a theoretical perspective showing that planning directly in action space suffers from an exponential decay in the probability of feasible trajectories as the horizon increases. 
In contrast, latent-space inference reshapes the search distribution toward feasible regions, enabling efficient long-horizon decision making. 
Extensive simulations and real-world experiments demonstrate that the WAV model consistently outperforms state-of-the-art methods, achieving significant improvements in task success rate, generalization ability, and robustness, especially in long-horizon and compositional scenarios.
\end{abstract}

\section{Introduction}

Building embodied agents that can robustly perceive, reason, and act in complex physical environments remains a central challenge in artificial intelligence~\citep{mason2001mechanics}. 
Vision--Language--Action (VLA) models have recently made significant progress by leveraging large-scale pre-trained vision--language models to map multimodal observations and natural language instructions to robot actions, enabling strong semantic generalization~\citep{brohan2022rt,zitkovich2023rt,kim2024openvla,black2024pi_0}. 
However, most existing VLA approaches rely on direct action prediction, treating each decision step independently and lacking mechanisms to reason over long-horizon trajectories or evaluate their outcomes. 
This limitation hinders their ability to perform coherent multi-step decision making in complex environments.

In parallel, world models aim to capture environment dynamics by predicting future observations, providing a structured representation of how the world evolves over time~\citep{wu2024ivideogpt,agarwal2025cosmos}. 
While such models enable reasoning about future states, they typically do not provide a principled mechanism for evaluating trajectories or selecting actions. 
Recent works attempt to unify world modeling and policy learning within a single architecture~\citep{cen2025worldvlaautoregressiveactionworld,liao2025genie,bi2025motusunifiedlatentaction,ye2026world,li2026causal}, but largely rely on supervised learning over static datasets, resulting in limited ability to capture fine-grained dynamics and assess long-horizon trajectory quality.
Another line of research leverages learned world models as simulators for data augmentation in RL~\citep{guo2025ctrl,li2025vla,zhu2025wmpo}. 
However, this paradigm is fundamentally constrained by the generalization capability of the world model, leading to compounding errors and unreliable simulated data in complex or out-of-distribution scenarios.
In contrast, model-based RL suggests that effective long-horizon decision making arises from the combination of predictive modeling and trajectory evaluation, where planning can be viewed as iteratively refining candidate futures based on their expected value~\citep{hansen2022temporal,hansen2023td,hansen2025learning}. 
This perspective raises a fundamental question: \emph{can planning be learned as an implicit inference process, rather than implemented as an explicit optimization module?}

In this work, we answer this question affirmatively by proposing \textbf{World--Value--Action (WAV\footnote{We refer to this model as WAV, emphasizing the WAVe-like propagation of value through predicted futures.})} model, a unified framework that endows VLA systems with implicit planning capabilities. 
Instead of performing explicit search in the high-dimensional action space, the WAV learns a latent distribution over future trajectories conditioned on visual observations and language instructions. 
A learned world model predicts future states, while a trajectory value function evaluates their long-horizon utility. 
Crucially, action generation is formulated as inference in this latent space, where the model progressively concentrates probability mass on trajectories that are both dynamically feasible and high-value.
We further provide a theoretical analysis showing that direct planning in action space suffers from an exponential decay in the probability of feasible trajectories as the planning horizon increases. 
In contrast, latent-space inference reshapes the search distribution toward feasible regions, while iterative refinement enables efficient optimization within this restricted domain. 
This perspective explains how planning can emerge implicitly from the joint learning of world dynamics and value estimation.
Our results demonstrate consistent improvements over strong baselines in task success rate, generalization, and robustness, with particularly large gains in long-horizon and compositional scenarios.
Our contributions are summarized as follows:
\begin{itemize}
\item We introduce \textbf{World--Value--Action (WAV)} model, a unified framework that enables implicit planning in VLA systems through joint modeling of dynamics and trajectory value.
\item We provide a theoretical perspective showing that latent planning mitigates the exponential decay of feasible trajectory probability inherent in action-space planning.
\item The WAV demonstrates state-of-the-art performance in simulated benchmarks and real-world scenarios, achieving significant improvements in long-horizon reasoning and generalization.
\end{itemize}

\section{Related Work}
\subsection{World Models for VLA}

World models have emerged as a core paradigm for equipping VLA models with predictive imagination and physical reasoning capabilities for robotic manipulation tasks. 
A large body of work focuses on unified architecture design, enabling joint modeling of policy execution and future state prediction within a shared transformer backbone, thereby bridging the gap between visual dynamics modeling and end-to-end robotic control in VLA systems~\cite{bi2025motus,cen2025rynnvla,BuQ-RSS-25,cen2025worldvla,zhang2025dreamvla,zhang2025gevrm,liao2025genie}. 
Subsequent efforts further improve the modeling capacity and policy--dynamics synergy of this paradigm, such as DreamVLA~\citep{zhang2025dreamvla}, GEVRM~\citep{zhang2025gevrm}, and Bu et al.~\citep{BuQ-RSS-25}, which enhance long-horizon prediction accuracy and manipulation performance within integrated VLA--world model systems.
Beyond architectural unification, another line of work explores world model-driven virtual environments for VLA post-training. 
NORA~\citep{hung2025nora} and RoboScape~\citep{tang2025roboscape} develop scalable pipelines for policy learning in simulation, while Xiao et al.~\citep{xiao2025world} and Li et al.~\citep{li2025vla} leverage world models to generate synthetic rewards for RL or preference optimization, enabling further policy refinement.
Rather than scaling model capacity or relying on external simulation, our work introduces implicit planning as an internal inference mechanism, offering a fundamentally different approach to long-horizon decision making in VLA systems.


\subsection{Model-Based Reinforcement Learning}
Model-based RL improves sample efficiency and long-horizon decision making by learning environment models for planning or policy optimization. 
Early work follows the Dyna paradigm and probabilistic dynamics modeling~\citep{sutton1991dyna,deisenroth2011pilco}, and is later extended with neural dynamics, uncertainty-aware model predictive control, and hybrid model-based (or model-free) optimization~\citep{chua2018deep,williams2018information,janner2019trust}. 
Latent world models further enable imagined rollouts from high-dimensional observations, as demonstrated in PlaNet, Dreamer, and MuZero~\citep{hafner2019learning,hafner2019dream,schrittwieser2020mastering}. 
More recent approaches emphasize scalability and task generalization, where methods such as TD-MPC and TD-MPC2~\citep{hansen2022temporal,hansen2023td} integrate latent planning with value learning, and subsequent work~\citep{hansen2025learning} shows that large-scale world models can generalize across diverse tasks and dynamics.
Rather than relying on explicit planning or iterative policy optimization, our approach treats planning as an emergent property of inference in a structured generative model, offering a fundamentally different perspective on long-horizon control.

\section{Preliminaries}
\paragraph{Vision--Language--Action (VLA) Models.}
We consider language-conditioned robotic manipulation, where at each timestep the model receives a visual observation $o_t \in \mathcal{O}$, a language instruction $g \in \mathcal{G}$, and optionally proprioceptive state $p_t \in \mathcal{P}$, and outputs an action $a_t \in \mathcal{A}$. Given an expert dataset $\mathcal{D}_{\text{expert}} = \{(g, \tau)\}$, where $\tau = \{(o_1, p_1, a_1), \dots, (o_T, p_T, a_T)\}$, a VLA model $\pi_\theta$ is modeled as a conditional distribution over actions, i.e., $\pi_\theta(a_{1:T} \mid o_{1:T}, p_{1:T}, g) = \prod_{t=1}^{T} \pi_\theta(a_t \mid o_{1:t}, p_{1:t}, a_{1:t-1}, g)$. In practice, the VLA model predicts short action sequences conditioned on the current context, $\pi_\theta(a_{t:t+H} \mid o_t, p_t, g)$, and is trained via maximum likelihood on $\mathcal{D}_{\text{expert}}$.

\paragraph{Model Predictive Control (MPC).}
MPC~\citep{garcia1989model} is a general paradigm for model-based decision making, where actions are selected by optimizing predicted future trajectories over a finite horizon.
At each decision step $t$, MPC solves:
\(
\textstyle
a_{t:t+H}^\star
=
\arg\max_{a_{t:t+H}}
\mathbb{E}
\left[
\sum_{i=0}^{H}
\gamma^{i}
R(s_{t+i}, a_{t+i})
\right]
\),
where $\gamma \in (0,1)$ is a discount factor and $R(\cdot)$ denotes the reward function.
Candidate trajectories are evaluated by rolling out a learned world model.
By explicitly reasoning over future trajectories, MPC achieves improved robustness and sample efficiency compared to purely reactive, model-free policies.

\section{Methodology}
In this work, we propose \textbf{WAV}, an end-to-end modular model for long-horizon reasoning and decision making via latent planning. 
We establish its theoretical foundation, instantiate it as a unified architecture, and realize it through iterative inference.

\subsection{Theoretical Analysis of Latent Planning}
\label{sec:theory}
\paragraph{Action-Space Planning and the Curse of Feasibility.} 
Long-horizon decision making in VLA tasks
poses a fundamental challenge for planning-based methods.
An intuitive idea of model-based planning seems to be to sample candidate action sequences and generalize them through a dynamic process.
However, as the planning horizon increases,
this approach quickly becomes ineffective.
The reason is not merely that the action space is high-dimensional,
but that the set of feasible trajectories occupies
an increasingly negligible fraction of the ambient trajectory space.
Formally, a horizon-$H$ trajectory resides in the space
$\mathcal{X} = \mathcal{S}^H \times \mathcal{A}^H$,
whose dimensionality grows linearly with $H$.
In VLA settings, visual constraints (e.g., scene geometry and object states), physical dynamics (e.g., contact and kinematics), and task-level semantic requirements (e.g., instruction grounding) jointly restrict valid behaviors to a thin manifold
$\mathcal{M}_{\mathrm{traj}} \subset \mathcal{X}$
with substantially lower intrinsic dimension.
Lemma~\ref{lem:vanishing_mass_main} formalizes the consequence of this mismatch: under uniform exploration in trajectory space, the probability of sampling even approximately feasible trajectories decays exponentially with the planning horizon.
This result reveals a fundamental limitation of direct action-space search in VLA systems: the vast majority of sampled trajectories are infeasible by construction, irrespective of the planner’s sophistication or the fidelity of the underlying models.

\begin{lemma}[Vanishing Feasible Mass in Trajectory Space]
\label{lem:vanishing_mass_main}
Let $\mathcal{X} = \mathcal{S}^H \times \mathcal{A}^H$ be a bounded trajectory space
with ambient dimension
$
D = H(\dim \mathcal{S} + \dim \mathcal{A}),
$
equipped with a reference measure $\mu$.
Let $\mathcal{M}_{\mathrm{traj}} \subset \mathcal{X}$ be a compact subset
with intrinsic dimension $d \ll D$.
Then, for any fixed tolerance $\epsilon \in (0,1)$,
there exists a constant $c > 0$ such that
\begin{equation}
\frac{\mu\big(\mathcal{N}_\epsilon(\mathcal{M}_{\mathrm{traj}})\big)}{\mu(\mathcal{X})}
\le \exp(-cH),
\end{equation}
where $\mathcal{N}_\epsilon(\mathcal{M}_{\mathrm{traj}})$ denotes the $\epsilon$-neighborhood
of $\mathcal{M}_{\mathrm{traj}}$.
\end{lemma}

\paragraph{Latent Planning as Probability Mass Reweighting.}
Latent planning alleviates the feasibility bottleneck in VLA tasks by reshaping the sampling distribution over trajectories.
Rather than sampling directly in the ambient action space, it samples a latent variable and decodes it into a full trajectory via a learned generative model.
When this generator captures the manifold of perceptually grounded and executable behaviors, latent sampling induces a distribution that concentrates probability mass on trajectories consistent with scene geometry, task semantics, and action affordances.
Proposition~\ref{prop:latent_reweighting_main} shows that, compared to uniform action-space sampling, this leads to exponentially higher probability of feasible trajectories as the horizon grows.
This advantage arises not from dimensionality reduction, but from reweighting the search toward dynamically and semantically valid regions, effectively acting as a probabilistic projection onto the feasible trajectory manifold.

\begin{proposition}[Latent Reweighting of the Search Distribution]
\label{prop:latent_reweighting_main}
Let $\mathcal{M}_{\mathrm{traj}} \subset \mathcal{X}$ denote the feasible trajectory set.
Consider a latent generator
$\mathcal{W}_\theta : \mathbb{R}^{d_z} \to \mathcal{X}$
and a state-conditioned latent distribution $f_\theta(s_t)$.
Assume that the learned model approximately parameterizes feasible trajectories
in the sense that there exists $\delta \in (0,1)$ such that
$
\textstyle
P_{\mathrm{latent}}(\mathcal{M}_{\mathrm{traj}})
=
\Pr_{z \sim f_\theta(s_t)}
\big[
\mathcal{W}_\theta(z) \in \mathcal{M}_{\mathrm{traj}}
\big]
\ge 1 - \delta .
$
Let $\Phi : \mathcal{A}^H \to \mathcal{X}$ denote the trajectory rollout map
induced by the system dynamics.
Then the ratio of feasible-trajectory probabilities satisfies
\begin{equation}
\frac{
\Pr_{z \sim f_\theta(s_t)}
\big[
\mathcal{W}_\theta(z) \in \mathcal{M}_{\mathrm{traj}}
\big]
}{
\Pr_{a_{t:t+H} \sim \mathrm{Unif}(\mathcal{A}^H)}
\big[
\Phi(a_{t:t+H}) \in \mathcal{M}_{\mathrm{traj}}
\big]
}
\ge
\exp(cH)\,(1-\delta).
\end{equation}
Consequently, latent planning assigns exponentially larger probability mass
to feasible trajectories relative to direct action-space sampling
as the horizon grows.
\end{proposition}

\paragraph{Why Feasibility Is Not Enough.}
While Proposition~\ref{prop:latent_reweighting} explains why latent planning avoids the exponential infeasibility of direct action-space search in VLA tasks, feasibility alone is insufficient for high-quality decision making.
The feasible set $\mathcal{M}_{\mathrm{traj}}$ still contains trajectories with widely varying task returns, from minimally valid behaviors to those that correctly satisfy language intent under visual constraints.
Thus, merely assigning non-vanishing probability mass to perceptually grounded and executable trajectories does not ensure that high-value solutions will be found with a practical number of samples.
Corollary~\ref{cor:iterative_inference_main} formalizes this limitation by showing that one-shot latent sampling fails under any subexponential sample budget.
Instead, effective VLA planning requires iterative inference that progressively concentrates probability mass within the feasible set toward trajectories that are both semantically aligned and high-value.

\begin{corollary}[Necessity of Iterative Latent Planning]
\label{cor:iterative_inference_main}
Let $\mathcal{M}_{\mathrm{traj}} \subset \mathcal{X}$ denote the feasible trajectory set,
and let
$
V(\tau)
=
\sum_{h=0}^{H-1} \gamma^h r(s_{t+h}, a_{t+h})
$
denote the cumulative discounted return of a trajectory $\tau$,
where $r$ is a bounded per-step reward and $\gamma \in (0,1]$.
Let
$\mathcal{W}_\theta : \mathbb{R}^{d_z} \to \mathcal{X}$
be a learned latent trajectory generator inducing the distribution
$P_{\mathrm{latent}} = (\mathcal{W}_\theta)_\# f_\theta$.
Assume that
$
P_{\mathrm{latent}}(\mathcal{M}_{\mathrm{traj}}) \ge 1 - \delta
\quad \text{for some } \delta \in (0,1).
$
Then, for any fixed sample budget $N$ that grows subexponentially with
the planning horizon $H$, one-shot sampling from $P_{\mathrm{latent}}$
does not, in general, guarantee with high probability
the discovery of a near-optimal trajectory.
Iterative inference procedures that adaptively reweight
the latent distribution are therefore necessary
to concentrate probability mass on high-value feasible trajectories.
\end{corollary}

Taken together, these results motivate the structure of proposed WAV model.
Latent planning resolves the feasibility bottleneck
by reshaping the search distribution,
while iterative inference resolves the remaining optimization challenge
within the feasible manifold.
Therefore, we first construct a modular end-to-end model architecture, and then explicitly combine these two components on this basis, thereby enabling efficient long-horizon planning in complex VLA tasks, as shown in Fig.~\ref{fig:pipline}.
The theoretical analysis details are given in Appendix Sec.~\ref{appendix:latent_planning}.
\begin{figure*}[htbp]
\centering
\includegraphics[width=0.95\columnwidth]{figures/vla_mpc_overview_1.png}
\caption{
The proposed WAV model decomposes planning and control into three tightly coupled components: 
\textbf{1)} a language-conditioned video generation module for predicting future visual trajectories; 
\textbf{2)} a trajectory value module for long-horizon evaluation; 
and \textbf{3)} an action module for generating executable robot actions. 
Together, these components enable implicit planning in latent space, where action generation arises from inference over predicted futures guided by trajectory value.
}
  \label{fig:pipline}
\end{figure*}

\subsection{Model Architecture and Training Objectives}
\paragraph{Language-Conditioned Video Generation Module.}
The video generation module predicts future visual trajectories conditioned on both past observations and task-level language instructions.
Given an instruction $g$, a frozen T5-XXL encoder~\citep{raffel2020exploring} produces a sequence of language embeddings
$\mathcal{T}(g) \in \mathbb{R}^{L_g \times d_t}$,
which are injected into a Diffusion Transformer (DiT)~\citep{peebles2023scalable} backbone via cross-attention.
Let $i \in \{h,l,r\}$ index camera views (e.g., head, left, right).
We denote by $\textstyle v_0^{(i)}$ the encoded initial observation, $\textstyle v_{\hat{t}}^{(i)}$ a sparse visual memory at time $\textstyle \hat{t}$, and $\textstyle z^{(i)} \sim \mathcal{N}(0,I)$ a view-specific latent noise variable.
The video generation module $\mathcal{W}$ predicts the next visual segment as
$\textstyle \hat{x}_{t:t+N} = \mathcal{W}\!\big(\{v_0^{(i)}, v_{\hat{t}}^{(i)}, z^{(i)}\}_{i}, \mathcal{T}(g)\big)$.
By jointly conditioning on multi-view visual history, stochastic latent variables, and language semantics, $\mathcal{W}$ captures both task intent and spatiotemporal dynamics.
Video generation proceeds autoregressively, yielding a temporally coherent visual rollout that represents the policy’s future execution.
Here, with a slight abuse of notation, we use $\mathcal{W}$
to denote the visual rollout component of the latent trajectory generator
$\mathcal{W}_\theta$ introduced in Section~\ref{sec:theory},
which produces the visual subspace of a full state--action trajectory.
\paragraph{Trajectory Value and Action Decoding Modules.}
Both the trajectory value and action decoding modules adopt a DiT-based autoregressive architecture operating on latent trajectory representations.
Visual features are first transformed into temporally structured tokens, which are then evaluated and decoded into actions.
At DiT block $i$, video tokens are processed as
$\mathbf{x}_i = \mathcal{B}_i^{\mathrm{vid}}\left(\mathbf{x}_{\mathrm{in}}, \mathcal{T}(g)\right)$,
where $\mathcal{B}_i^{\mathrm{vid}}$ denotes the $i^{th}$ visual transformer block.
The trajectory value module initializes a latent value token
$\mathbf{z}_{\mathrm{val}} \sim \mathcal{N}(0,I)$
and refines it through cross-attention with video features:
$\mathbf{u}_i = \mathcal{B}_i^{\mathrm{val}}\big(\mathbf{z}_{\mathrm{val}}, \operatorname{CrossAttn}(\mathbf{z}_{\mathrm{val}}, \mathbf{x}_i)\big)$,
where $\mathbf{u}_i$ represents the latent trajectory value embedding.
The action decoder initializes latent action tokens $\mathbf{z}_{\mathrm{act}}$ and sequentially integrates video and value feature:
{
\begin{equation}
\begin{aligned}
\mathbf{z}_{\mathrm{act}}^{(i)}
&=
\mathcal{B}_i^{\mathrm{act}}\!\left(
\mathbf{z}_{\mathrm{act}}^{(i-1)},
\operatorname{CrossAttn}(\mathbf{z}_{\mathrm{act}}^{(i-1)}, \mathbf{x}_i)
\right), \text{and } 
\mathbf{a}_i
=
\mathcal{B}_i^{\mathrm{act}}\!\left(
\mathbf{z}_{\mathrm{act}}^{(i)},
\operatorname{CrossAttn}(\mathbf{z}_{\mathrm{act}}^{(i)}, \mathbf{u}_i)
\right).
\end{aligned}
\label{eq:action}
\end{equation}
}
\paragraph{Training via Flow Matching.}
We adopt a three-stage training strategy based on flow matching.
Following standard flow matching formulations,
we assume that $x^0, v^0, a^0$ are samples drawn from the corresponding base noise distributions
for video, trajectory return, and robot actions, respectively,
while $x^1, v^1, a^1$ denote samples from the target data distributions,
including the target video trajectories,
the return
$v^1 = \sum_{i=0}^{H} \gamma^i R(s_{t+i}, a_{t+i})$,
and the target robot action distribution.
Here we use a rule-based dense reward function similar to that used in previous work--ReinboT~\citep{zhang2025reinbot} (details in Appendix Sec.~\ref{appendix:Implementation_Details}).
First, the video generation module is trained using a video flow loss:
$\mathcal{L}_{\mathrm{vid}} = \mathbb{E}\!\left[\| v_\theta(t, l, o, x^t) - (x^1 - x^0) \|_2^2 \right]$.
The video model is then frozen, and a trajectory value module is trained to predict cumulative discounted rewards:
$\mathcal{L}_{\mathrm{val}} = \mathbb{E}\big[\| v_\theta(t, l, o, z_{\mathrm{vid}}, v^t) - (v^1 - v^0) \|_2^2 \big]$.
Finally, all modules are jointly trained with an action flow loss:
$\mathcal{L}_{\mathrm{act}} = \mathbb{E}\big[\| v_\theta(t, l, o, z_{\mathrm{vid}}, z_{\mathrm{val}}, a^t) - (a^1 - a^0) \|_2^2 \big]$.

\begin{wrapfigure}{r}{0.6\columnwidth}
\vspace{-10pt}
\begin{minipage}{0.56\columnwidth}
\begin{algorithm}[H]
\caption{WAV: Latent Trajectory Planning}
\label{alg:latent_mppi}
\begin{algorithmic}[1]
\State Video generator module $\mathcal{W}$, trajectory value module $\mathcal{V}$, initial flow noise distributions $f_{\mathrm{vid}}^{(0)}, f_{\mathrm{val}}^{(0)}$, samples $(M,N)$, elites $(K_1,K_2)$.
\For{iteration $k = 1,\dots,K$}
    \State Sample video latent noise $\mathbf{z}_{\mathrm{vid}}^{(m)} \sim f_{\mathrm{vid}}^{(k-1)}$
    \State Generate video feature $\mathbf{x}^{(m)}$ 
    \For{each video feature $\mathbf{x}^{(m)}$}
        \State Sample value latent noise $\mathbf{z}_{\mathrm{val}}^{(m,n)} \sim f_{\mathrm{val}}^{(k-1)}$
        \State Evaluate values $\mathbf{U}^{(m,n)}$ 
        \State Compute score $\phi^{(m)}$ 
    \EndFor
    \State Update $f_{\mathrm{vid}}^{(k)}$ using top-$K_1$ samples by Eq.~\ref{eq:video_mu_sigma}
    \State Update $f_{\mathrm{val}}^{(k)}$ using top-$K_2$ samples by Eq.~\ref{eq:value_mu_sigma}
\EndFor
\State Sample optimized latent noise $\mathbf{z}_{\mathrm{vid}}^\star, \mathbf{z}_{\mathrm{val}}^\star$
\State Predict action by Eq.~\ref{eq:action}
\end{algorithmic}
\end{algorithm}
\end{minipage}
\vspace{-10pt}
\end{wrapfigure}

\subsection{Latent Planning and Iterative Inference}
Inspired by Model Predictive Path Integral (MPPI)~\citep{williams2015model} control, we perform iterative latent-space trajectory optimization during inference.
Instead of fixing latent priors to standard Gaussians, we adaptively update the noise distributions of the video and value modules based on trajectory evaluations.
We maintain two independent Gaussian noise distributions:
a video flow noise distribution
$
\scriptsize
f_{\mathrm{vid}}^{(k)} = \mathcal{N}\!\left(
\boldsymbol{\mu}_{\mathrm{vid}}^{(k)},
\operatorname{diag}\!\left((\boldsymbol{\sigma}_{\mathrm{vid}}^{(k)})^2\right)
\right),
$
and a trajectory value flow noise distribution
$
\scriptsize
f_{\mathrm{val}}^{(k)} = \mathcal{N}\!\left(
\boldsymbol{\mu}_{\mathrm{val}}^{(k)},
\operatorname{diag}\!\left((\boldsymbol{\sigma}_{\mathrm{val}}^{(k)})^2\right)
\right),
$
where $k \in \{1,\dots,K\}$ denotes the iteration index.

At iteration $k$, we first sample $M$ groups of video latent noise:
$
\scriptstyle
\left\{
\mathbf{z}_{\mathrm{vid}}^{(m)}
\right\}_{m=1}^M
\sim
f_{\mathrm{vid}}^{(k-1)}.
$
Each latent sample is denoised by the video generation module $\mathcal{W}$ to produce video feature chunking:
$
\scriptstyle
\mathbf{x}^{(m)}_{1:H_1} = \mathcal{W}\!\big(\mathbf{z}_{\mathrm{vid}}^{(m)}\big)$,
where $\mathbf{x}^{(m)} \in \mathbb{R}^{H_1 \times d_v}$ represents a predicted future visual trajectory feature.
For each video feature chunking $\mathbf{x}^{(m)}$, we sample $N$ groups of trajectory value flow noise:
$
\scriptstyle
\left\{
\mathbf{z}_{\mathrm{val}}^{(m,n)}
\right\}_{n=1}^N
\sim
f_{\mathrm{val}}^{(k-1)},
$
and denoise them using the trajectory value module $\mathcal{V}$ conditioned on the same video features:
$
\scriptstyle
\mathbf{v}^{(m,n)} = \mathcal{V}\!\big(\mathbf{x}^{(m)}, \mathbf{z}_{\mathrm{val}}^{(m,n)}\big)$,
where $\mathbf{v}^{(m,n)}$ denotes a predicted state-value.

To obtain a stable assessment of future trajectory quality, we compute the Signal-to-Noise Ratio (SNR) of the value predictions for each exploration:
$\scriptstyle \operatorname{SNR}^{(m,n)} = \mathbb{E}\!\left[\mathbf{v}^{(m,n)}\right] / \left(\operatorname{Std}\!\left[\mathbf{v}^{(m,n)}\right] + \epsilon\right)$
where $\epsilon$ is a small constant for numerical stability.
The final score of the $m^{th}$ video exploration is defined as the most reliable evaluation among its $N$ value estimates:
$
\scriptstyle
\phi^{(m)} = \max_{n \in \{1,\dots,N\}} \operatorname{SNR}^{(m,n)}$.
Based on the scores $\{\phi^{(m)}\}_{m=1}^M$, we select the top $K_1$ elite video explorations.
The video noise distribution is updated using the empirical mean and standard deviation of their corresponding latent variables:
{
\scriptsize
\begin{equation}
\begin{aligned}
\boldsymbol{\mu}_{\mathrm{vid}}^{(k)}
&=
\frac{1}{K_1}
\sum_{m \in \mathcal{E}_{\mathrm{vid}}}
\mathbf{z}_{\mathrm{vid}}^{(m)}, \text{and }
\boldsymbol{\sigma}_{\mathrm{vid}}^{(k)}
=
\sqrt{
\frac{1}{K_1}
\sum_{m \in \mathcal{E}_{\mathrm{vid}}}
\left(
\mathbf{z}_{\mathrm{vid}}^{(m)} -
\boldsymbol{\mu}_{\mathrm{vid}}^{(k)}
\right)^2
}.
\end{aligned}
\label{eq:video_mu_sigma}
\end{equation}
}
Similarly, from all $M \times N$ value evaluations, we select the top $K_2$ samples with the highest SNR and update the value noise distribution:
{\scriptsize
\begin{equation}
\begin{aligned}
\boldsymbol{\mu}_{\mathrm{val}}^{(k)}
&=
\frac{1}{K_2}
\sum_{(m,n) \in \mathcal{E}_{\mathrm{val}}}
\mathbf{z}_{\mathrm{val}}^{(m,n)},
 \text{and }
\boldsymbol{\sigma}_{\mathrm{val}}^{(k)}
=
\sqrt{
\frac{1}{K_2}
\sum_{(m,n) \in \mathcal{E}_{\mathrm{val}}}
\left(
\mathbf{z}_{\mathrm{val}}^{(m,n)} -
\boldsymbol{\mu}_{\mathrm{val}}^{(k)}
\right)^2
}.
\end{aligned}
\label{eq:value_mu_sigma}
\end{equation}
}

To prevent distribution collapse and improve iteration stability, we apply variance decay and exponential smoothing:
\(
\scriptstyle
\boldsymbol{\mu}^{(k)} \!\leftarrow\!
\alpha \boldsymbol{\mu}^{(k)} + (1-\alpha)\boldsymbol{\mu}^{(k-1)},\;
\boldsymbol{\sigma}^{(k)} \!\leftarrow\!
\beta \boldsymbol{\sigma}^{(k)} + (1-\beta)\boldsymbol{\sigma}^{(k-1)}
\)
with smoothing parameters $\alpha, \beta \in (0,1)$.
After $K$ iterations, we sample latent variables from the optimized distributions
$f_{\mathrm{vid}}^{(K)}$ and $f_{\mathrm{val}}^{(K)}$,
and obtain the corresponding video and value features via flow denoising.
These features are finally fused by the action decoding module to predict decision-making actions.
Algo.~\ref{alg:latent_mppi} summarizes the proposed latent planning procedure.

\section{Experiments}
In this section, we comprehensively explore how the proposed WAV model effectively implements latent trajectory planning, thereby improving the performance of visual language manipulation in general scenarios. 
To this end, our experiments aim to investigate the following questions: 
~\textbf{1)} Compared with baseline algorithms, does WAV exhibit better generalization ability and higher success rate when performing robot manipulation tasks? 
~\textbf{2)} Can WAV effectively generalize to complex long-horizon real-world manipulation tasks? 
~\textbf{3)} How important is latent trajectory planning based on the world model to the overall generalization performance of WAV model? 
~\textbf{4)} What are the inference efficiency and memory overhead of WAV model?

\subsection{Generalization Comparison in Simulation}

\paragraph{Simulation setup.} 
We evaluate the WAV on the LIBERO benchmark~\citep{liu2023libero}, a comprehensive suite for multi-task and lifelong robotic manipulation.
LIBERO consists of four task suites designed to probe different aspects of generalization.
\textbf{Spatial} evaluates spatial reasoning by varying scene layouts while keeping object identities fixed.
\textbf{Object} tests object-level generalization by changing object instances within a fixed environment.
\textbf{Goal} assesses goal-conditioned behavior by varying task goals under the same scene configuration.
Finally, \textbf{Long} includes long-horizon compositional tasks involving diverse objects, layouts, and goals, posing significant challenges for temporal consistency and multi-step reasoning.
A single WAV model is deployed across four suites.
\begin{table}[htbp]
\caption{Performance comparison on the LIBERO benchmark.}
\label{tab:libero_only}
\centering
\small
\resizebox{0.8\columnwidth}{!}{
\begin{tabular}{lcccccc}
\toprule
\multirow{2}{*}{\centering\textbf{Models}} 
& \multirow{2}{*}{\textbf{Params}}
& \multicolumn{4}{c}{\textbf{LIBERO}} 
& \multirow{2}{*}{\textbf{Avg.}} \\
\cmidrule(lr){3-6}
& & \textbf{Spatial} & \textbf{Object} & \textbf{Goal} & \textbf{Long} & \\
\midrule
Diffusion Policy \citet{chi2023diffusion} & - & 78.3 & 92.5 & 68.3 & 50.5 & 72.4 \\
Octo \citet{team2024octo} & - & 78.9 & 85.7 & 84.6 & 51.1 & 75.1 \\
OpenVLA \citet{kim2024openvla} & 7b & 84.9 & 88.4 & 79.2 & 53.7 & 76.5 \\
SpatialVLA \citet{qu2025spatialvla} & 4b & 88.2 & 89.9 & 78.6 & 55.5 & 78.1 \\
DiT Policy \citet{hou2024diffusion} & - & 84.2 & 96.3 & 85.4 & 63.8 & 82.4 \\
$\pi_{0}$-FAST \citet{pertsch2025fast} & 3b & 96.4 & 96.8 & 88.6 & 60.2 & 85.5 \\
GR00T-N1 \citet{bjorck2025gr00t} & 2b & 94.4 & 97.6 & 93.0 & 90.6 & 93.9 \\
$\pi_{0}$ \citet{black2024pi_0} & 3b & 96.8 & 98.8 & 95.8 & 85.2 & 94.2 \\
VLA-0 \citet{goyal2025vla} & 3b & 97.0 & 97.8 & 96.2 & 87.6 & 94.7 \\
Discrete Diffusion VLA \citet{liang2025discrete} & 7b & 97.2 & 98.6 & 97.4 & 92.0 & 96.3 \\
MemoryVLA \citet{shi2025memoryvla} & 7b & 98.4 & 98.4 & 96.4 & 93.4 & 96.5 \\
$\pi_{0.5}$ \citet{intelligence2025pi_} & 3b & 98.8 & 98.2 & 98.0 & 92.4 & 96.8 \\
OpenVLA-OFT \citet{kim2025fine} & 7b & 97.6 & 98.4 & 97.9 & 94.5 & 97.1 \\
VLA-Adapter \citet{wang2025vla} & 0.5b & 97.8 & 99.2 & 97.2 & \textbf{95.0} & 97.3 \\
\midrule
\rowcolor{gray!15}
\multicolumn{7}{l}{\textit{World Model}} \\
WorldVLA \citet{cen2025worldvla} & 7b & 85.6 & 89.0 & 82.6 & 59.0 & 79.1 \\
CoT-VLA \citet{zhao2025cot} & 7b & 87.5 & 91.6 & 87.6 & 69.0 & 81.1 \\
FlowVLA \citet{zhong2025flowvla} & 8.5b & 93.2 & 95.0 & 91.6 & 72.6 & 88.1 \\
DreamVLA \citet{zhang2025dreamvla} & 3b & 97.5 & 94.0 & 89.5 & 89.5 & 92.6 \\
UD-VLA \citet{chen2025unified} & - & 94.1 & 95.7 & 91.2 & 89.6 & 92.7 \\
RynnVLA-002-Discrete \citet{cen2025rynnvla} & 7b & 94.2 & 96.8 & 94.6 & 87.6 & 93.3 \\
UniVLA \citet{bu2025univla} & 7b & 95.4 & 98.8 & 93.5 & 94.0 & 95.5 \\
$\mathcal{F}_1$\citet{lv2025f1} & 4.5b & 98.2 & 97.8 & 95.4 & 91.3 & 95.7 \\
GE-ACT \citet{liao2025genie} & 2b & 98.2 & 97.6 & 95.8 & 94.4 & 96.5 \\
\midrule
\rowcolor{lightblue}
\textbf{WAV (Ours)} & 2.2b & \textbf{99.6} & \textbf{100.0} & \textbf{98.6} & 94.4 & \textbf{98.1} \\
\qquad W/o Latent Trajectory Planning &  & 99.0 & 99.6 & 95.0 & 91.8 & 96.4 \\
\bottomrule
\end{tabular}
}
\end{table}

\paragraph{Results.} 
Tab.~\ref{tab:libero_only} reports results on the LIBERO benchmark. 
Our method achieves state-of-the-art (or competitive) performance across all task suites, with an overall average score of 98.1. 
These results highlight its effectiveness in handling long-horizon and compositional tasks, where efficient trajectory-level inference is critical.
To assess the contribution of implicit planning, we perform an ablation by removing latent trajectory planning. 
This leads to an average performance drop of 1.7 points, with the largest degradation on the \textbf{Long} suite, decreasing from 94.4 to 91.8. 
This trend underscores the importance of trajectory-level planning for robust and generalizable task execution.
Notably, the performance gap widens with increasing task horizon, suggesting that implicit planning is particularly beneficial for mitigating compounding errors in long-horizon scenarios.

\subsection{Real-World Evaluation}
\paragraph{Real-world setup.} 
We evaluate WAV on a real-world dual-arm robotic platform, Piper, across a set of challenging manipulation tasks designed to probe key policy capabilities, including spatial reasoning, deformable object manipulation, visual grounding, and long-horizon planning. 
Representative tasks include \textbf{bowl organization}, \textbf{towel flattening}, and a \textbf{long-horizon drawer task} that requires opening a drawer, placing an object inside, and closing it.
For a fair comparison, we select GE-ACT~\citep{liao2025genie} as the baseline model, as it uses the same video pre-trained model and has a similar network architecture and parameter count.
Following standard practice, we adopt a strict binary success metric: a trial is considered successful only if the task is fully completed, with no partial credit. 
More details are in Appendix Sec.~\ref{appendix:Implementation_Details}.

\begin{wrapfigure}{r}{0.5\textwidth}
    \centering
    \includegraphics[width=1\linewidth]{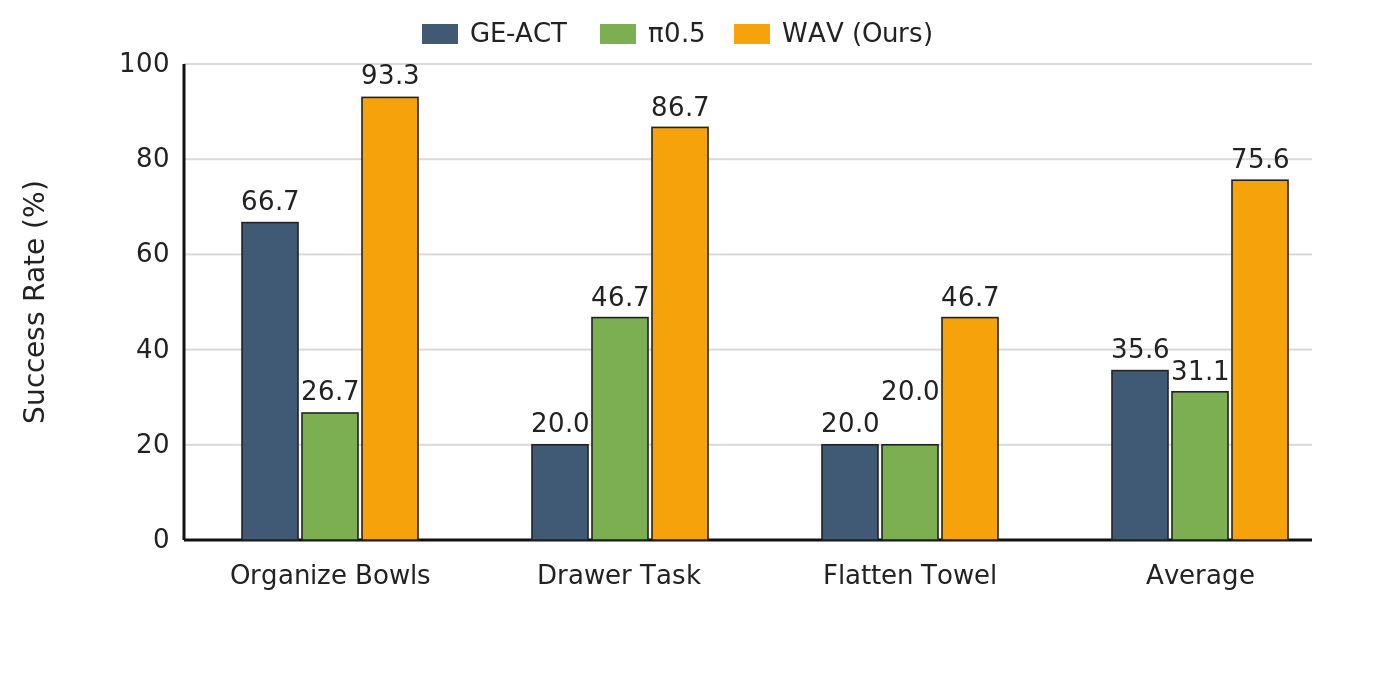}
    \caption{
    Quantitative comparison of WAV with GE-ACT and $\pi_{0.5}$ on real-world tasks.
    Each result is averaged over 15 trials.
    }
    \label{fig:real_world_bar}
\end{wrapfigure}

\begin{figure}[htbp]
\centering
\includegraphics[width=\textwidth]{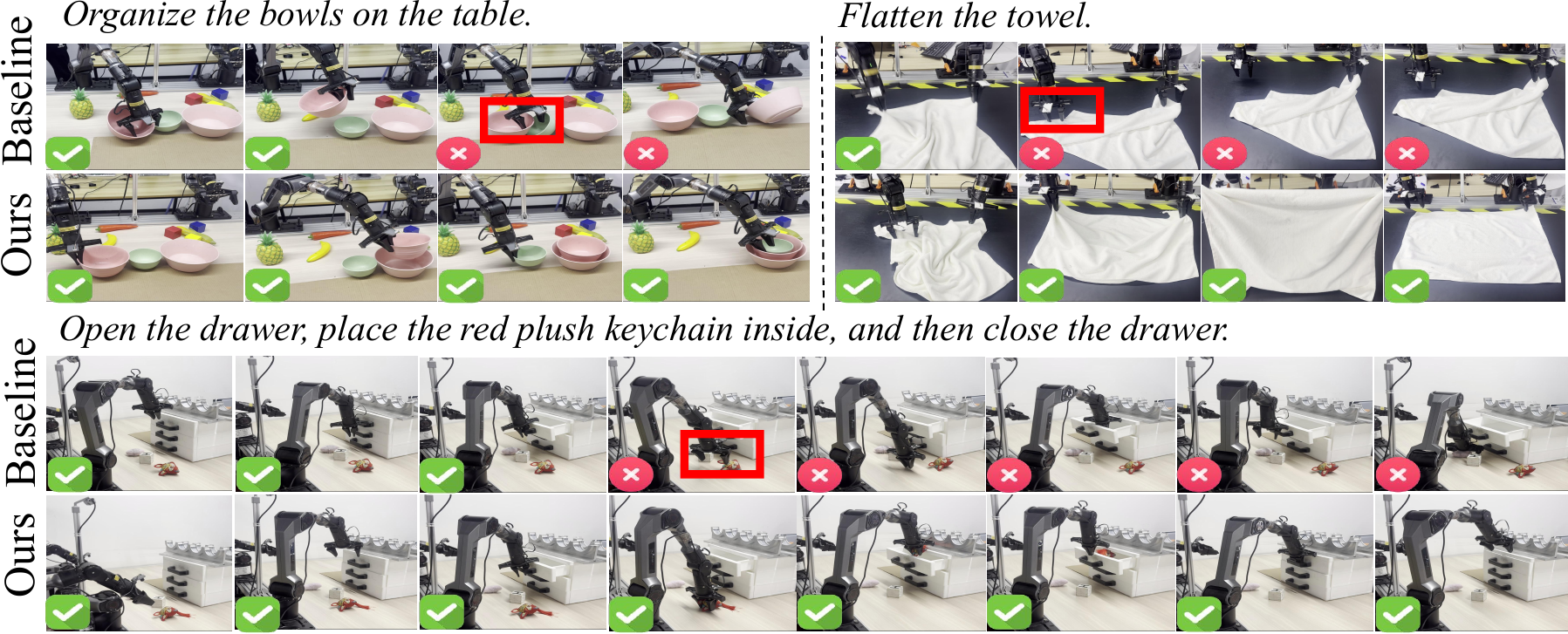}
\caption{
Qualitative comparison between WAV model (Ours) and the GE-ACT (Baseline) across different real-world manipulation tasks. 
}
  \label{fig:real_world_tasks}
\end{figure}

\paragraph{Results.} 
As shown in Fig.~\ref{fig:real_world_bar}, WAV achieved a significantly higher success rate across all tasks than the baseline, with the average success rate increasing from 35.6\% to 75.6\%.
Qualitative results (Fig.~\ref{fig:real_world_tasks}) reveal distinct failure modes of the baseline. 
GE-ACT frequently exhibits inaccurate action execution and weak spatial grounding, such as misaligning with the drawer handle or failing to grasp objects reliably. 
These errors accumulate over multiple steps, leading to cascading failures in long-horizon tasks. 
In contrast, WAV produces more consistent and coherent multi-step behaviors. 
By implicitly planning over future trajectories, it anticipates downstream consequences of actions, resulting in more accurate interactions and improved robustness. 
This advantage is most pronounced in tasks requiring multi-step coordination, highlighting the role of implicit planning in mitigating error accumulation in real-world settings.


\subsection{Ablation Study}
\begin{figure}
\centering
\includegraphics[width=0.8\textwidth]{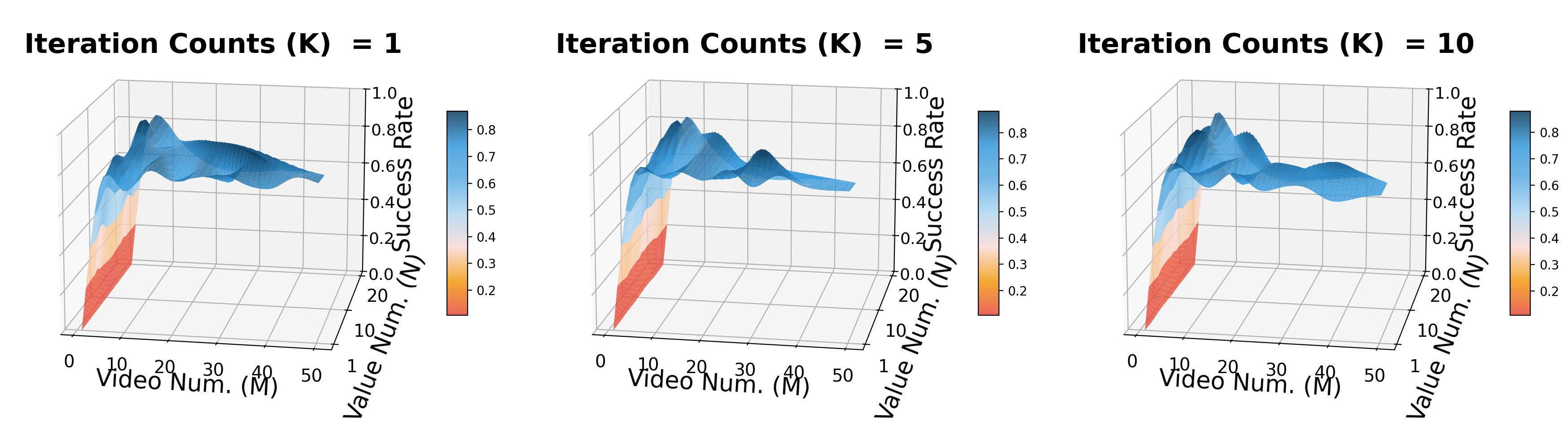}
\caption{
WAV performance under varying iteration counts $K$, video latent noise sampling numbers $M$, and trajectory value noise sampling numbers $N$.
}
  \label{fig:stage1_main}
\end{figure}
\paragraph{Effect of Iteration Counts $K$ and Sampling Sizes $M, N$.}
As shown in Fig.~\ref{fig:stage1_main}, 
increasing $K$ from 1 to 5 leads to a clear improvement in success rate, indicating that iterative latent planning effectively refines the trajectory distribution toward high-value regions. 
However, further increasing $K$ to 10 yields only marginal gains, suggesting diminishing returns from excessive iterations. 
In contrast, the performance is highly sensitive to $M$, where increasing the number of video samples results in a steep improvement, especially in the low-sample regime. 
This highlights the importance of sufficiently exploring future trajectory hypotheses. 
The influence of $N$ is comparatively milder, with performance saturating earlier, indicating that dense value evaluation provides limited additional benefit once a reasonable estimation is achieved.
More results are given in Appendix Fig.~\ref{app_fig:stage1}.
Appendix Fig.~\ref{app_fig:value} and~\ref{app_fig:gt_pred_libero} compare trajectory value and video predictions to ground truth, respectively.

\begin{wrapfigure}{r}{0.53\textwidth}
\centering
\includegraphics[width=0.95\linewidth]{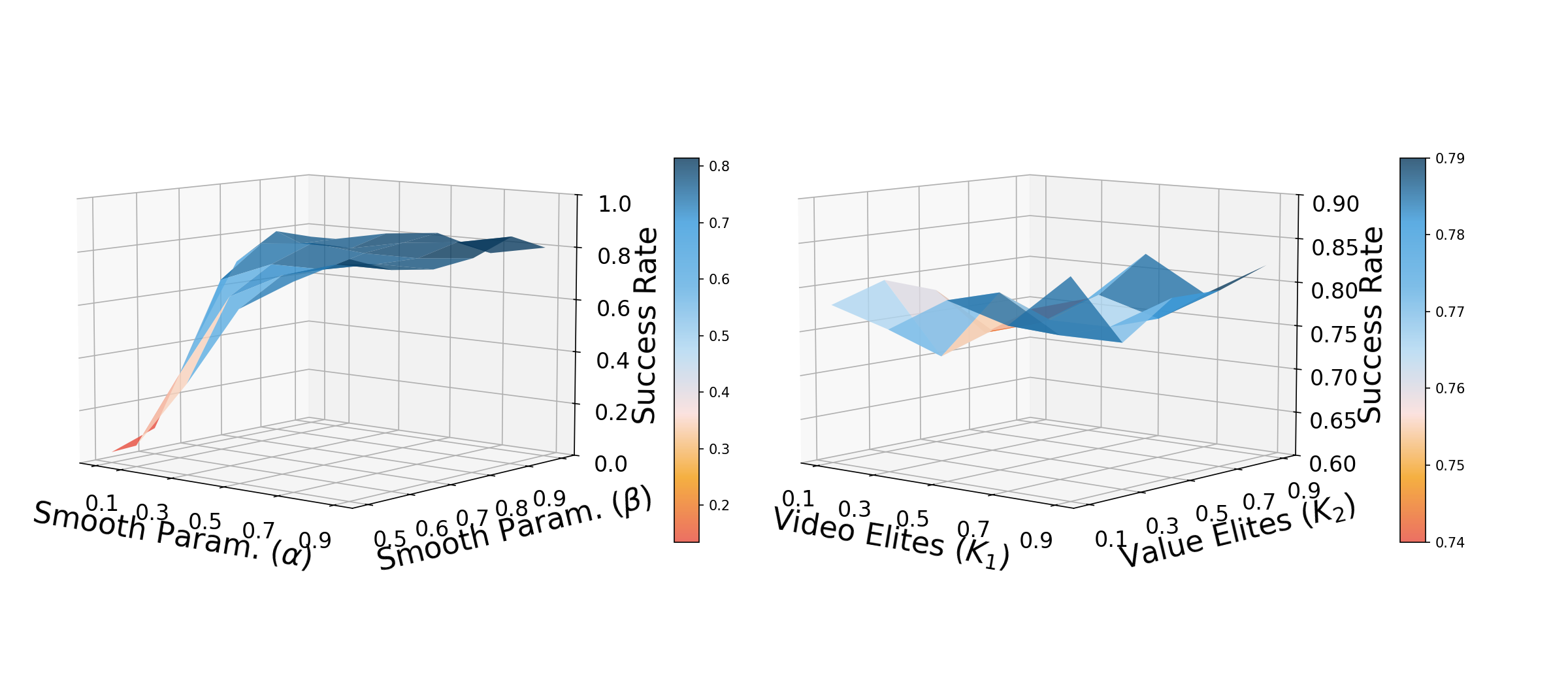}
\caption{
\textbf{Left:} WAV sensitivity to smoothing parameters $\alpha$ and $\beta$.
\textbf{Right:} WAV sensitivity to elite counts $K_1$ and $K_2$.
}
  \label{fig:stage2_stage3_vis}
\end{wrapfigure}
\paragraph{Effect of Smoothing Parameters ($\alpha, \beta$).}
Fig.~\ref{fig:stage2_stage3_vis} (\textbf{Left}) shows that when $\alpha$ is small, the success rate drops sharply, indicating that insufficient smoothing leads to unstable distribution updates and poor convergence. 
As $\alpha$ increases, performance improves rapidly and then plateaus, showing a clear steep region followed by saturation. A similar but slightly milder trend is observed for $\beta$. This suggests that stabilizing the update of the latent distribution is crucial, but overly strong smoothing provides limited additional benefit once convergence is achieved.

\paragraph{Effect of Elite Selection ($K_1, K_2$).}
Fig.~\ref{fig:stage2_stage3_vis} (\textbf{Right}) shows that the performance is relatively stable across different choices of $K_1$ and $K_2$, indicating that the WAV is not highly sensitive to these parameters. 
However, extremely small values lead to noticeable performance degradation, suggesting that a sufficient number of elite samples is necessary for stable distribution updates.


\begin{wrapfigure}{r}{0.53\textwidth}
\centering
\includegraphics[width=0.95\linewidth]{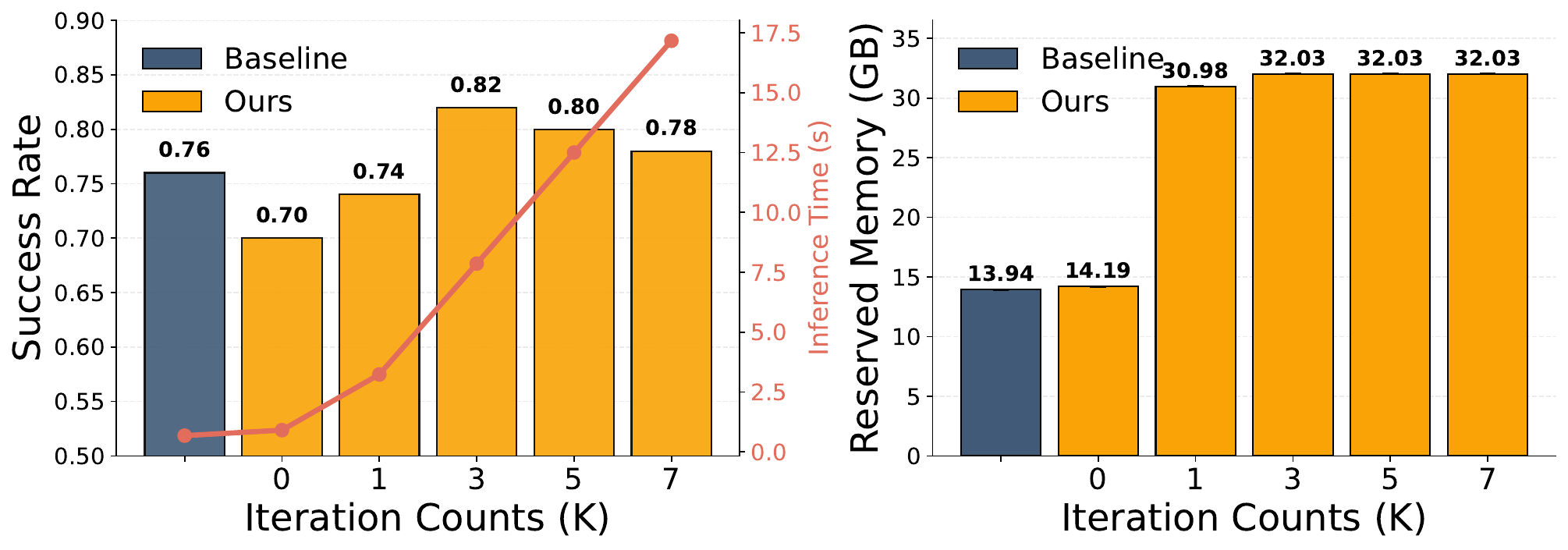}
\caption{
Performance and efficiency trade-off of WAV across $K$: success rate and inference time (\textbf{Left}), and GPU memory usage (\textbf{Right}).
Baseline: GE-ACT.
Ours: WAV.
}
\label{fig:speed_acc}
\end{wrapfigure}
\paragraph{Performance--Efficiency Trade-off.} 
Fig.~\ref{fig:speed_acc} shows that increasing the number of iterations from $0$ to $3$ consistently improves performance. 
Beyond $3$ gains saturate while computational cost, particularly inference time, continues to grow. 
Therefore, $K=3$ provides a favorable balance between effectiveness and efficiency, achieving strong performance with moderate computational overhead.
This behavior reflects the diminishing returns of iterative refinement, indicating that a small number of inference steps is sufficient to capture most of the planning benefits.



\section{Conclusion}
In this work, we introduced \textbf{WAV}, a unified framework for integrating perception, reasoning, and action in embodied agents. 
The key innovation lies in latent trajectory planning, which maintains a compact representation of plausible futures and reshapes the search distribution toward high-value, low-uncertainty regions, thereby significantly reducing the complexity of long-horizon planning. 
Empirical results demonstrate that WAV consistently improves task success, generalization, and robustness over previous baselines across simulation and real-world settings.
The main limitation of current work is the time and storage overhead required for model deployment.
Future directions include extending the framework to richer multi-modal instructions and enabling real-time closed-loop deployment on physical robotic systems.
More broadly, our results suggest that planning can be naturally realized as an inference process within structured generative models, offering a promising direction for unifying prediction, evaluation, and control.

\section*{Acknowledgments}
This work was supported by the Brain Science and Brain-like Intelligence Technology --- National Science and Technology Major Project (Grant No. 2022ZD0208800).


\bibliography{neurips_2026}
\bibliographystyle{plainnat}


\newpage
\appendix
\onecolumn

\section{Theoretical Analysis of Latent Planning}
\label{appendix:latent_planning}

In this section, we provide a principled explanation for why latent
planning is substantially more effective than direct action-space sampling
in long-horizon Vision--Language--Action (VLA) tasks.
Our analysis is not intended as a tight optimality guarantee,
but rather as a geometric and probabilistic characterization of the search
difficulty inherent to high-dimensional trajectory spaces, and how latent
planning alleviates it by reweighting probability mass toward feasible
trajectories.

\subsection{Setup}
\label{app:theory_setup}

Let the environment be a Markov decision process
$(\mathcal{S}, \mathcal{A}, \mathcal{P}, R, \gamma)$.
A planning horizon-$H$ trajectory lies in the space
\[
\mathcal{X} = \mathcal{S}^H \times \mathcal{A}^H,
\]
whose ambient dimension grows linearly with $H$.
We denote by $\mathcal{M}_{\mathrm{traj}} \subset \mathcal{X}$
the set of trajectories that satisfy physical dynamics, contact constraints,
and semantic task requirements.
Throughout, we treat $\mathcal{M}_{\mathrm{traj}}$ as a thin subset of
$\mathcal{X}$ with low intrinsic dimension, consistent with the manifold
hypothesis commonly assumed in robotics and control.

We consider a learned latent trajectory generator
$\mathcal{W}_\theta : \mathbb{R}^{d_z} \rightarrow \mathcal{X}$,
which maps a latent variable $z$ to a full trajectory,
\[
\tau_{t:t+H} = \mathcal{W}_\theta(z),
\qquad
z \sim f_\theta(s_t),
\]
where $f_\theta(s_t)$ is a state-conditioned latent distribution.
The induced trajectory distribution is the pushforward measure
\[
P_{\mathrm{latent}} = (\mathcal{W}_\theta)_\# f_\theta .
\]

\subsection{Vanishing Feasible Mass in Trajectory Space}
\label{app:vanishing_mass}

We first formalize the difficulty of direct action-space sampling
by characterizing how the probability mass of feasible trajectories
shrinks with the planning horizon.

\begin{lemma}[Vanishing Feasible Mass in Trajectory Space]
\label{lem:vanishing_mass}
Let $\mathcal{X} = \mathcal{S}^H \times \mathcal{A}^H$ be a bounded
trajectory space with ambient dimension
\[
D = H(\dim \mathcal{S} + \dim \mathcal{A}),
\]
equipped with a reference measure $\mu$.
Let $\mathcal{M}_{\mathrm{traj}} \subset \mathcal{X}$ be a compact subset
with intrinsic dimension $d \ll D$.
Then, for any fixed tolerance $\epsilon \in (0,1)$,
there exists a constant $c > 0$ such that
\begin{equation}
\frac{
\mu\big(\mathcal{N}_\epsilon(\mathcal{M}_{\mathrm{traj}})\big)
}{
\mu(\mathcal{X})
}
\leq
\exp(-cH),
\end{equation}
where $\mathcal{N}_\epsilon(\mathcal{M}_{\mathrm{traj}})$ denotes the
$\epsilon$-neighborhood of $\mathcal{M}_{\mathrm{traj}}$.
\end{lemma}

\begin{proof}
We work in the bounded trajectory space
$\mathcal{X} \subset \mathbb{R}^D$
equipped with the Lebesgue measure $\mu$.
Let $\mathcal{M}_{\mathrm{traj}} \subset \mathcal{X}$ be compact
with intrinsic dimension $d$.

Fix any $\epsilon \in (0,1)$ and define
\[
\mathcal{N}_\epsilon(\mathcal{M}_{\mathrm{traj}})
=
\left\{
x \in \mathcal{X}
\mid
\inf_{y \in \mathcal{M}_{\mathrm{traj}}}
\|x-y\|_2 \leq \epsilon
\right\}.
\]

By standard covering number arguments for compact sets,
$\mathcal{M}_{\mathrm{traj}}$ can be covered by at most
$N_\epsilon \leq C_1 \epsilon^{-d}$ Euclidean balls of radius $\epsilon$,
where $C_1 > 0$ is independent of $H$.
Thus,
\[
\mathcal{N}_\epsilon(\mathcal{M}_{\mathrm{traj}})
\subset
\bigcup_{i=1}^{N_\epsilon} B(y_i,2\epsilon).
\]

Each such ball has $D$-dimensional volume proportional to $\epsilon^D$.
Up to multiplicative factors that grow at most subexponentially with $D$,
we obtain
\[
\mu(\mathcal{N}_\epsilon(\mathcal{M}_{\mathrm{traj}}))
\lesssim
\epsilon^{D-d}.
\]

Since $\mathcal{X}$ is bounded, $\mu(\mathcal{X}) < \infty$.
Moreover, the ambient dimension grows linearly with the horizon,
\[
D = H(\dim \mathcal{S} + \dim \mathcal{A}),
\]
while the intrinsic dimension of feasible trajectories is governed by
physical and semantic constraints.
We assume that
\[
d \leq \lambda H
\quad
\text{for some }
\lambda < \dim \mathcal{S} + \dim \mathcal{A},
\]
so that there exists $\kappa > 0$ satisfying
$D-d \geq \kappa H$.

Since $\log \epsilon < 0$, it follows that
\[
\epsilon^{D-d}
=
\exp\!\big((D-d)\log \epsilon\big)
\leq
\exp\!\big(-\kappa H |\log \epsilon|\big).
\]

Absorbing constant and subexponential terms into the exponent,
there exists $c>0$ such that
\[
\frac{
\mu(\mathcal{N}_\epsilon(\mathcal{M}_{\mathrm{traj}}))
}{
\mu(\mathcal{X})
}
\leq
\exp(-cH),
\]
which completes the proof.
\end{proof}

\paragraph{Discussion.}
Lemma~\ref{lem:vanishing_mass} states that,
under uniform exploration of the trajectory space,
the probability of encountering even an approximately feasible trajectory
decays exponentially with the planning horizon.
This phenomenon arises from the mismatch between the rapidly growing ambient
dimension and the low intrinsic dimensionality of feasible behaviors,
and captures a fundamental scaling challenge faced by long-horizon
action-space planning.

\subsection{Latent Planning as Probability Mass Reweighting}
\label{app:latent_reweighting}

We now contrast action-space sampling with latent-space planning.
Rather than comparing absolute volumes,
we compare the probability mass assigned to feasible trajectories
under different induced distributions.

\begin{proposition}[Latent Reweighting of the Search Distribution]
\label{prop:latent_reweighting}
Let $\mathcal{M}_{\mathrm{traj}} \subset \mathcal{X}$ denote the feasible
trajectory set.
Consider a latent generator
$\mathcal{W}_\theta : \mathbb{R}^{d_z} \to \mathcal{X}$
and a state-conditioned latent distribution $f_\theta(s_t)$.
Assume that the learned model approximately parameterizes feasible
trajectories in the sense that there exists $\delta \in (0,1)$ such that
\begin{equation}
P_{\mathrm{latent}}(\mathcal{M}_{\mathrm{traj}})
=
\Pr_{z \sim f_\theta(s_t)}
\left[
\mathcal{W}_\theta(z)
\in
\mathcal{M}_{\mathrm{traj}}
\right]
\geq 1-\delta.
\end{equation}

Let $\Phi : \mathcal{A}^H \to \mathcal{X}$ denote the trajectory rollout
map induced by the system dynamics.
Then the ratio of feasible-trajectory probabilities satisfies
\begin{equation}
\frac{
\Pr_{z \sim f_\theta(s_t)}
\left[
\mathcal{W}_\theta(z)
\in
\mathcal{M}_{\mathrm{traj}}
\right]
}{
\Pr_{a_{t:t+H} \sim \mathrm{Unif}(\mathcal{A}^H)}
\left[
\Phi(a_{t:t+H})
\in
\mathcal{M}_{\mathrm{traj}}
\right]
}
\geq
\exp(cH)(1-\delta),
\end{equation}
where $c>0$ is the constant from Lemma~\ref{lem:vanishing_mass}.
Consequently, latent planning assigns exponentially larger probability mass
to feasible trajectories relative to direct action-space sampling
as the horizon grows.
\end{proposition}

\begin{proof}
Uniform sampling over $\mathcal{A}^H$ induces a trajectory distribution
via the rollout map $\Phi$.
By Lemma~\ref{lem:vanishing_mass},
the probability of sampling an $\epsilon$-feasible trajectory satisfies
\[
\Pr_{a_{t:t+H} \sim \mathrm{Unif}(\mathcal{A}^H)}
\left[
\Phi(a_{t:t+H})
\in
\mathcal{M}_{\mathrm{traj}}
\right]
\leq
\exp(-cH).
\]

On the other hand, latent sampling induces the pushforward distribution
\[
P_{\mathrm{latent}}
=
(\mathcal{W}_\theta)_\# f_\theta.
\]
By assumption,
\[
\Pr_{z \sim f_\theta(s_t)}
\left[
\mathcal{W}_\theta(z)
\in
\mathcal{M}_{\mathrm{traj}}
\right]
\geq
1-\delta.
\]

Taking the ratio of the two probabilities yields the stated result.
\end{proof}

\paragraph{Discussion.}
Proposition~\ref{prop:latent_reweighting} is a conditional comparison:
given a learned latent representation that approximately captures
the feasible trajectory set, latent planning avoids the exponential decay
of probability mass that plagues action-space search.
This advantage arises from reweighting the search distribution toward
dynamically and semantically valid regions,
rather than from exhaustive exploration of a lower-dimensional space.

Proposition~\ref{prop:latent_reweighting} establishes that latent planning
reweights the search distribution toward the feasible trajectory manifold,
thereby mitigating the exponential infeasibility of action-space exploration.
Nevertheless, feasibility alone is insufficient for long-horizon planning,
as the feasible set still contains trajectories with widely varying returns.
The following corollary formalizes the necessity of iterative latent planning.

\subsection{Necessity of Iterative Latent Planning}
\label{app:iterative_inference}

\begin{corollary}[Necessity of Iterative Latent Planning]
\label{cor:iterative_inference}
Let $\mathcal{M}_{\mathrm{traj}} \subset \mathcal{X}$ denote the feasible
trajectory set, and let
\[
V(\tau)
=
\sum_{h=0}^{H-1}
\gamma^h r(s_{t+h},a_{t+h})
\]
denote the cumulative discounted return of a trajectory $\tau$,
where $r$ is a bounded per-step reward and $\gamma \in (0,1]$.
Let
$\mathcal{W}_\theta : \mathbb{R}^{d_z} \to \mathcal{X}$
be a learned latent trajectory generator inducing the distribution
$P_{\mathrm{latent}} = (\mathcal{W}_\theta)_\# f_\theta$.

Assume that
\[
P_{\mathrm{latent}}(\mathcal{M}_{\mathrm{traj}})
\geq
1-\delta
\quad
\text{for some }
\delta \in (0,1).
\]

Then, for any fixed sample budget $N$ that grows subexponentially with
the planning horizon $H$, one-shot sampling from
$P_{\mathrm{latent}}$ does not, in general, guarantee with high probability
the discovery of a near-optimal trajectory.
Iterative inference procedures that adaptively reweight
the latent distribution are therefore necessary
to concentrate probability mass on high-value feasible trajectories.
\end{corollary}

\begin{proof}
Define the optimal feasible trajectory value
\[
V^\star
=
\sup_{\tau \in \mathcal{M}_{\mathrm{traj}}}
V(\tau).
\]
For any $\varepsilon > 0$, define the $\varepsilon$-optimal feasible set
\[
\mathcal{M}_\varepsilon
=
\left\{
\tau \in \mathcal{M}_{\mathrm{traj}}
\mid
V(\tau) \geq V^\star-\varepsilon
\right\}.
\]

By construction, $\mathcal{M}_\varepsilon$ is a measurable subset of
$\mathcal{M}_{\mathrm{traj}}$.
While the latent distribution assigns non-vanishing probability mass
to the feasible set $\mathcal{M}_{\mathrm{traj}}$,
this assumption alone does not imply any uniform lower bound on
$P_{\mathrm{latent}}(\mathcal{M}_\varepsilon)$.
In particular, the probability mass of $P_{\mathrm{latent}}$
may be dispersed across trajectories with widely varying values.

Let $\tau^{(1)}, \ldots, \tau^{(N)}$ be $N$ independent samples
drawn from $P_{\mathrm{latent}}$.
The probability of observing at least one $\varepsilon$-optimal trajectory is
\[
\Pr
\left[
\exists i \leq N:
\tau^{(i)} \in \mathcal{M}_\varepsilon
\right]
=
1-
\left(
1-P_{\mathrm{latent}}(\mathcal{M}_\varepsilon)
\right)^N.
\]

If $N$ grows subexponentially with $H$,
then unless $P_{\mathrm{latent}}(\mathcal{M}_\varepsilon)$
is bounded below by a constant independent of $H$,
the above probability cannot be guaranteed to remain bounded away from zero
as the planning horizon increases.
Therefore, without additional mechanisms that adaptively reweight
the latent distribution toward higher-value regions,
one-shot latent sampling does not provide a reliable strategy
for long-horizon optimization.

Iterative inference methods, such as importance-weighted resampling,
cross-entropy optimization, or gradient-based refinement in latent space,
explicitly modify the sampling distribution based on observed returns,
thereby increasing the probability mass assigned to
$\mathcal{M}_\varepsilon$ over successive iterations.
This establishes the necessity of iterative latent planning.
\end{proof}

\section{Additional Experimental Results}
\label{app:additional_results}

This section provides additional experiments that further analyze the
source of the performance gains, the robustness of the value-guided
planning mechanism, and the inference-time computation--performance
trade-off.

\subsection{Attribution of Performance Gains}
\label{app:attribution}

To isolate the source of the performance gains, we conduct a controlled
attribution study on LIBERO. All variants use the same frozen video,
value, and action modules, and differ only in their inference-time
strategies. Each adjacent comparison changes a single component,
allowing us to separately evaluate the effects of additional sampling,
value-based selection, and iterative refinement.

\begin{table}[htbp]
\centering
\scriptsize
\setlength{\tabcolsep}{4pt}
\renewcommand{\arraystretch}{1.08}
\begin{tabular}{lcccccccc}
\toprule
\textbf{Inference strategy} &
\textbf{Value} &
\textbf{Iter.} &
\textbf{Fwd.} &
\textbf{Spatial} &
\textbf{Object} &
\textbf{Goal} &
\textbf{Long} &
\textbf{Avg.}
\\
\midrule

One-shot sampling ($K=0$) &
-- & -- & $1\times$ &
99.0 & 99.6 & 95.0 & 91.8 & 96.4
\\

Random selection &
-- & -- & $M\times$ &
99.0 & 99.0 & 96.8 & 92.2 & 96.7
\\

Best-of-$N$, mean-value selection &
\checkmark & -- & $M\times$ &
99.4 & 99.4 & 96.4 & 93.0 & 97.0
\\

Best-of-$N$, max-SNR selection &
\checkmark & -- & $M\times$ &
99.0 & 98.8 & 97.6 & 93.6 & 97.2
\\

Compute-matched Best-of-$N$ &
\checkmark & -- & $KM\times$ &
99.0 & 98.8 & 97.6 & 93.6 & 97.2
\\

WAV w/o value conditioning &
-- & \checkmark & $KM\times$ &
99.6 & 99.0 & 96.0 & 93.8 & 97.1
\\

\textbf{WAV (full, $K=3$)} &
\checkmark & \checkmark & $KM\times$ &
\textbf{99.6} & \textbf{100.0} & \textbf{98.6} &
\textbf{94.4} & \textbf{98.1}
\\

\bottomrule
\end{tabular}
\caption{
Attribution study of WAV on LIBERO.
All variants use identical frozen video, value, and action modules;
only the inference strategy differs.
}
\label{tab:attribution}
\end{table}

The attribution ladder reveals four observations.
First, additional sampling alone provides little benefit:
random selection improves the Long suite by only 0.4 points over
one-shot sampling (92.2 vs.~91.8).
Second, learned trajectory values provide a meaningful selection signal,
with both mean-value and max-SNR selection outperforming random selection.
The max-SNR criterion further improves Long and Goal over mean-value
selection.

Third, the gain of WAV cannot be explained by Best-of-$N$ or additional
sampling alone. Under the same number of forward passes, WAV improves
over compute-matched Best-of-$N$ by 0.8 points on Long
and 1.0 points on Goal.
Moreover, increasing the one-shot candidate pool from $M$ to $KM$
does not improve performance.

Finally, removing value conditioning from iterative refinement reduces
Goal performance from 98.6 to 96.0 while retaining the iterative
procedure. This isolates the contribution of value guidance from the
iteration mechanism itself. Overall, WAV improves the average score
from 96.4 to 98.1, with a larger gain of 2.6 points on the Long suite.

\subsection{Robustness to Reward Design}
\label{app:reward_robustness}

We further investigate whether WAV depends critically on the particular
dense reward formulation used to train the value module.
We retrain only the value module using a sparse binary success reward
while keeping all other components unchanged.
The dense reward uses only RGB observations and proprioceptive signals
available on the robot, as described in Sec.~\ref{app:reward_construction}.

\begin{table}[htbp]
\centering
\scriptsize
\setlength{\tabcolsep}{5pt}
\renewcommand{\arraystretch}{1.08}
\begin{tabular}{lcccccc}
\toprule
\textbf{Value training reward} &
\textbf{Signal source} &
\textbf{Spatial} &
\textbf{Object} &
\textbf{Goal} &
\textbf{Long} &
\textbf{Avg.}
\\
\midrule
Rule-based dense &
RGB + proprioception &
99.6 & 100.0 & 98.6 & 94.4 & 98.1
\\
Sparse success only &
Task outcome &
98.8 & 99.4 & 98.0 & 97.4 & 98.4
\\
w/o planning &
-- &
99.0 & 99.6 & 95.0 & 91.8 & 96.4
\\
\bottomrule
\end{tabular}
\caption{
Robustness to reward design on LIBERO.
Only the value module is retrained; all other components remain unchanged.
}
\label{tab:reward_ablation}
\end{table}

The sparse-reward variant achieves 98.4 average performance, comparable
to the dense-reward variant (98.1) and above the no-planning reference
(96.4). This indicates that the effectiveness of value-guided iterative
planning is not tied to the specific dense reward formulation.
The dense reward provides higher performance on Spatial and Object,
whereas the sparse variant performs better on Long.
Overall, both reward designs retain a clear advantage over the
no-planning reference.

\subsection{Inference Cost and Accuracy--Compute Trade-off}
\label{app:inference_cost}

WAV exposes $K$ as a test-time computation--performance trade-off.
For each planning call, the model predicts an action chunk of eight
steps and executes the chunk before the next planning call.
Thus, the effective control frequency is
$8/T$, where $T$ is the planning latency.

\begin{table}[htbp]
\centering
\scriptsize
\setlength{\tabcolsep}{4pt}
\renewcommand{\arraystretch}{1.08}
\begin{tabular}{cccccccc}
\toprule
\textbf{$K$} &
\textbf{$M$} &
\textbf{$N$} &
\textbf{Mean Lat. (s)} &
\textbf{P95 Lat. (s)} &
\textbf{Freq. (Hz)} &
\textbf{Peak Mem. (GB)} &
\textbf{LIBERO-10 Avg.}
\\
\midrule
0 & 1 & 1 &
0.909 & 0.970 & 8.80 & 14.2 & 70
\\
1 & 30 & 5 &
3.237 & 3.274 & 2.47 & 31.0 & 74
\\
3 & 30 & 5 &
7.862 & 7.888 & 1.02 & 32.0 & \textbf{82}
\\
5 & 30 & 5 &
12.500 & 12.526 & 0.64 & 32.0 & 80
\\
\bottomrule
\end{tabular}
\caption{
Inference latency, memory, effective control frequency, and accuracy
under different numbers of iterative updates $K$ on a single
NVIDIA A100 80\,GB GPU.
Each planning call predicts and executes an eight-step action chunk.
}
\label{tab:inference_cost}
\end{table}

For example, at $K=3$, the measured latency is 7.86\,s,
corresponding to an effective control frequency of approximately
1.02\,Hz. On the harder LIBERO-10 subset, increasing $K$ from 0 to 3
improves the average performance from 70 to 82, while increasing $K$
further to 5 reduces performance to 80.
This demonstrates that iterative refinement is not monotonically
beneficial and that $K$ serves as a test-time scaling parameter rather
than a fixed computational cost.

The measured peak GPU memory remains approximately 32\,GB for
$K\geq1$, while latency increases with the number of iterations.
These results characterize the practical regime targeted by WAV:
quasi-static, long-horizon manipulation rather than high-frequency
reactive control.

\subsection{Sensitivity to Inference Hyperparameters}
\label{app:sensitivity}

We further report the sensitivity of WAV to the number of iterative
updates $K$, video latent samples $M$, and trajectory value noise
samples $N$. The results characterize the interaction between sampling
and iterative refinement and provide additional guidance for selecting
the inference-time configuration.

\begin{figure}[htbp]
\centering
\includegraphics[width=0.95\textwidth]{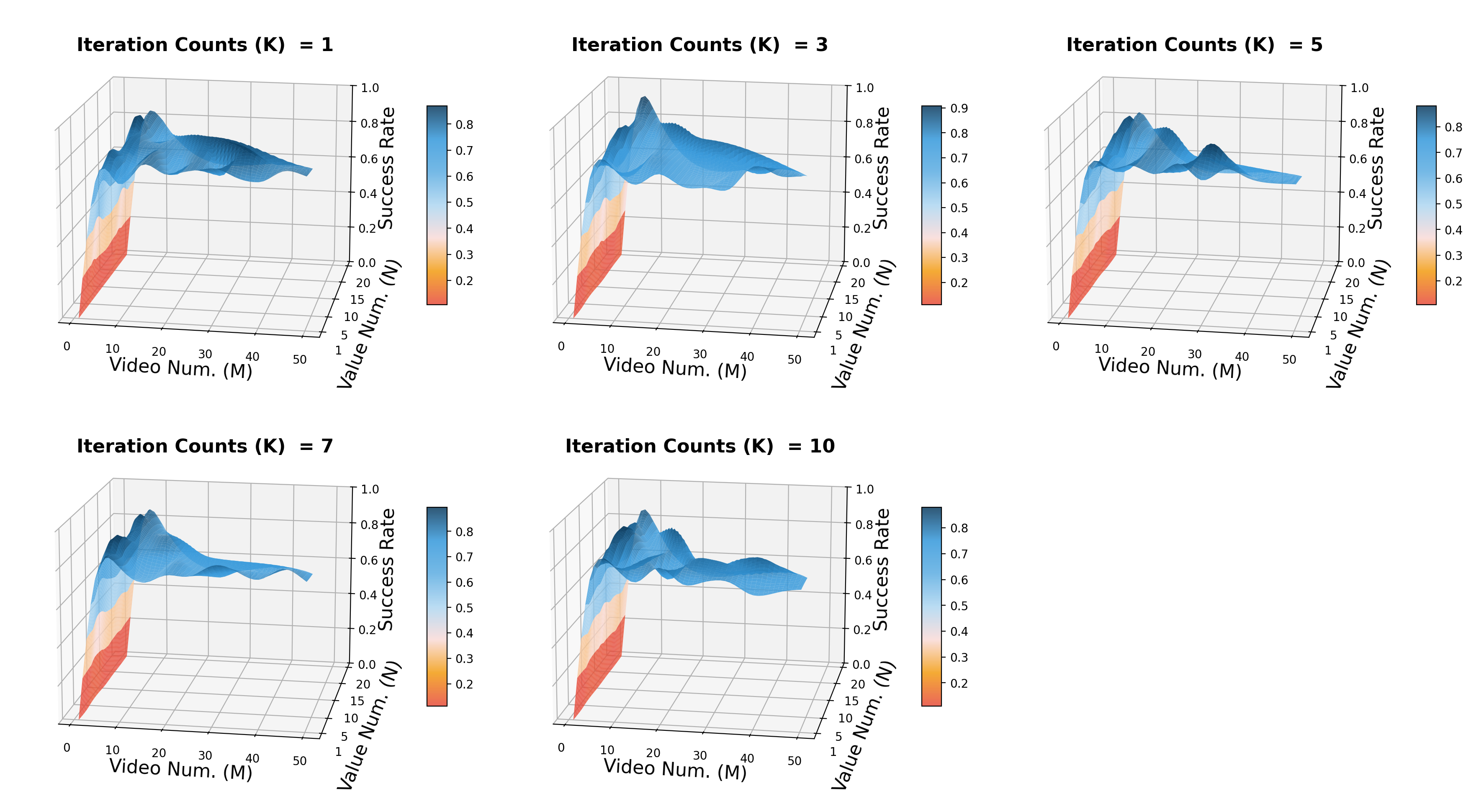}
\caption{
Performance variation trends of the WAV model under different iteration
counts $K$, video latent noise sampling numbers $M$, and trajectory
value noise sampling numbers $N$.
}
\label{app_fig:stage1}
\end{figure}

\subsection{Value Prediction}
\label{app:value_prediction}

We compare the inferred state-value trajectories with their corresponding
ground-truth values in both real-world robot experiments and LIBERO.
The value estimates capture the overall temporal trend of trajectory
quality across both environments.

\begin{figure}[htbp]
\centering
\includegraphics[width=0.95\textwidth]{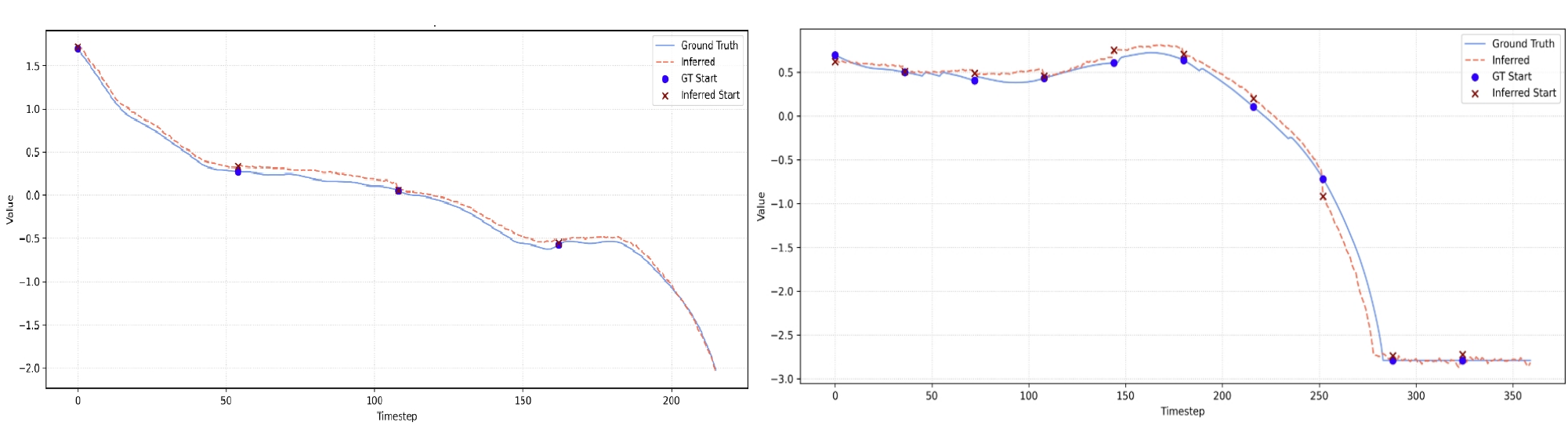}
\caption{
Comparison between predicted and ground-truth state-value trajectories
in real-world robot experiments (left) and LIBERO (right).
}
\label{app_fig:value}
\end{figure}

\subsection{Qualitative Video Comparison}
\label{app:qualitative}

\begin{figure}[htbp]
\centering
\includegraphics[width=0.95\textwidth]{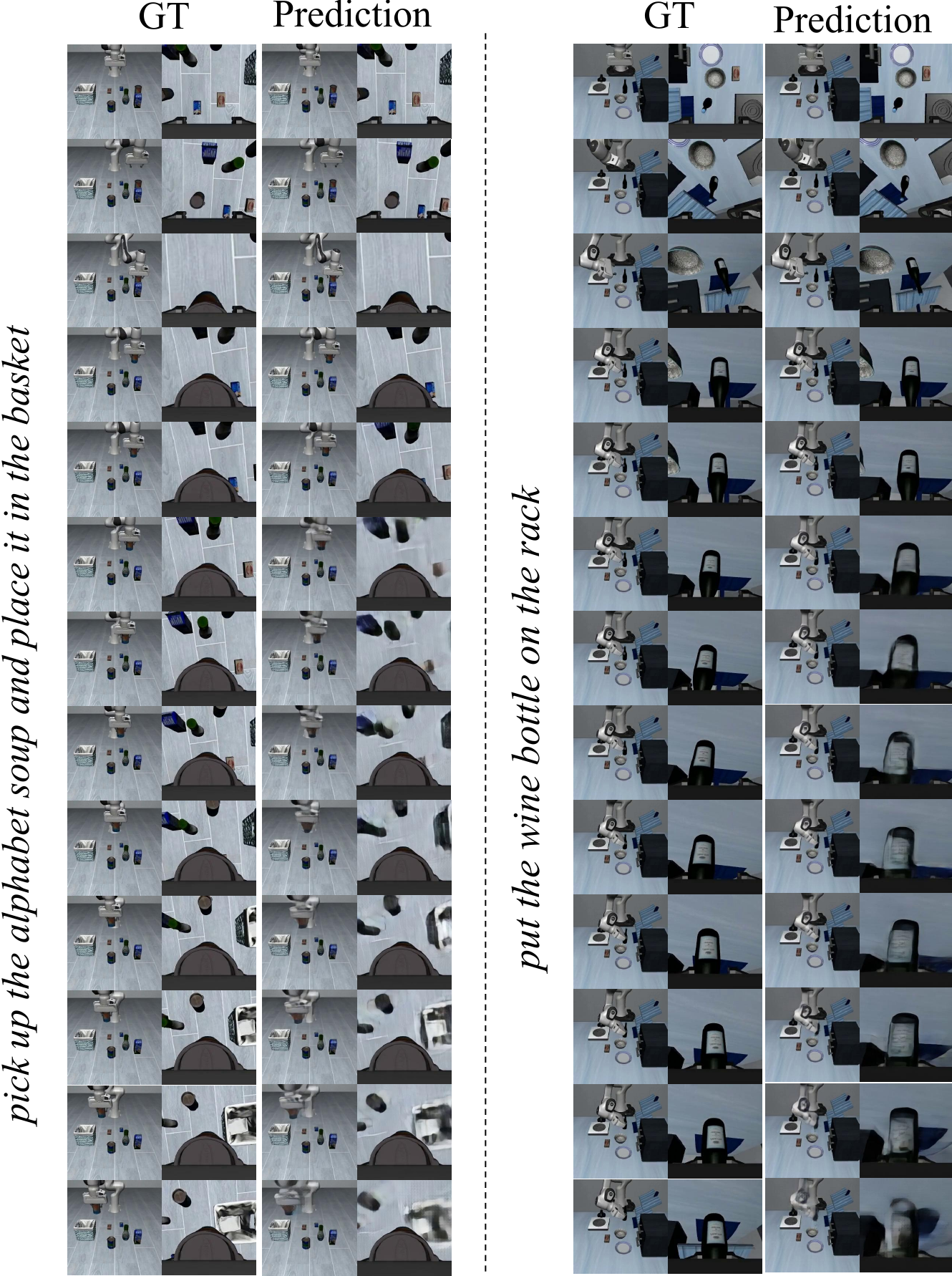}
\caption{
Qualitative comparison between predicted videos and ground-truth
observations on two LIBERO tasks.
}
\label{app_fig:gt_pred_libero}
\end{figure}

We additionally compare the videos generated by the learned video model
with the corresponding ground-truth observations on representative
LIBERO tasks. The results are shown in Figure ~\ref{app_fig:gt_pred_libero} 

\section{Implementation Details}
\label{appendix:Implementation_Details}

\subsection{Real-world Data Collection}
\label{app:data_collection}

To evaluate the WAV model in real-world settings, we collected
approximately 300 successful trajectories each for the drawer-opening
and bowl-organization tasks, and 2,000 successful trajectories for the
towel-flattening task, using a real-world Piper robot.

Each trajectory lasts on the order of tens of seconds.
Each trajectory contains RGB observations from two wrist-mounted cameras
with resolution $240\times320\times3$ and one third-person top camera
with resolution $240\times424\times3$, together with 14-dimensional
robot joint states.

We train one task-specific model per task and evaluate each model on
the corresponding real-world scenario.
This setup enables a controlled and systematic evaluation across diverse
manipulation challenges, while highlighting the robustness of the
proposed WAV model in real-world deployment.

\subsection{Reward Construction}
\label{app:reward_construction}

For dense reward design, we follow ReinboT~\citep{zhang2025reinbot},
which decomposes dense rewards into four aspects:
sub-goal achievement, task progress, behavior smoothness,
and task completion.
In this work, we omit the sub-goal achievement term and retain the
remaining components.

Concretely, our reward consists of nine reward terms, corresponding to
16 scalar components in the bimanual setting.
The definitions and weights are listed in Tab.~\ref{tab:rewards}.
The dataset we use will be publicly available upon publication.

\begin{table}[htbp]
\centering
\scriptsize
\setlength{\tabcolsep}{3pt}
\renewcommand{\arraystretch}{1.08}
\begin{tabularx}{\columnwidth}{p{2.9cm}Yp{2cm}}
\hline
Reward term & Definition & Weight \\
\hline

Wrist-view MSE reward
&
$c_{1,t}^b=
\exp\!\left(
-0.01\cdot\mathrm{MSE}(I_t^b,I_T^b)
\right)$
&
$+\frac{1}{16}$ each
\\

Wrist-view SSIM reward
&
$c_{2,t}^b=
\exp\!\left(
\mathrm{SSIM}(I_t^b,I_T^b)-1
\right)$
&
$+\frac{1}{16}$ each
\\

Top-view MSE reward
&
$c_{3,t}=
\exp\!\left(
-0.01\cdot\mathrm{MSE}
(I_t^{\mathrm{top}},I_T^{\mathrm{top}})
\right)$
&
$+\frac{1}{16}$
\\

Top-view SSIM reward
&
$c_{4,t}=
\exp\!\left(
\mathrm{SSIM}
(I_t^{\mathrm{top}},I_T^{\mathrm{top}})-1
\right)$
&
$+\frac{1}{16}$
\\

Joint-state proximity reward
&
$c_{5,t}^b=
\exp\!\left(
-\|s_t^b-s_T^b\|_2
\right)$
&
$+\frac{1}{16}$ each
\\

Joint velocity penalty
&
$c_{6,t}^b=
\sum_j|\Delta s_{t,j}^b|$
&
$-\frac{1}{16}$ each
\\

Joint acceleration penalty
&
$c_{7,t}^b=
\sum_j|\Delta^2s_{t,j}^b|$
&
$-\frac{1}{16}$ each
\\

Action velocity penalty
&
$c_{8,t}^b=
\sum_j|\Delta a_{t,j}^b|$
&
$-\frac{0.1}{16}$ each
\\

Action acceleration penalty
&
$c_{9,t}^b=
\sum_j|\Delta^2a_{t,j}^b|$
&
$-\frac{0.1}{16}$ each
\\

\hline
\end{tabularx}
\caption{
Compact formulation of the dense reward terms in the bimanual setting,
where $b\in\{L,R\}$ denotes the left and right arms.
}
\label{tab:rewards}
\end{table}

\subsection{Training Configuration}
\label{app:training}

The hyperparameter configurations for video training and value/action
training are summarized in Tables~\ref{tab:trainer1}
and~\ref{tab:trainer2}.

\begin{table}[htbp]
\centering
\caption{Video training hyperparameter configuration.}
\begin{tabular}{cc}
\toprule
\textbf{Parameter} & \textbf{Value} \\
\hline
Gradient clip & 1.0 \\
Steps & 40000 \\
Warm-up steps & 1000 \\
Batch size & 128 \\
Learning rate & $3e-4$ \\
Weight decay & $1e-5$ \\
Caption Dropout & 0.06 \\
Optimizer &
Adam ($\beta_1=0.9,\ \beta_2=0.95,\ \beta_3=0.999$) \\
\bottomrule
\end{tabular}
\label{tab:trainer1}
\end{table}

\begin{table}[htbp]
\centering
\caption{Value and action training hyperparameter configuration.}
\begin{tabular}{cc}
\toprule
\textbf{Parameter} & \textbf{Value} \\
\hline
Gradient clip & 1.0 \\
Steps & 30000 \\
Warm-up steps & 1000 \\
Batch size & 128 \\
Learning rate & $5e-5$ \\
Weight decay & $1e-5$ \\
Caption Dropout & 0 \\
Optimizer &
Adam ($\beta_1=0.9,\ \beta_2=0.95,\ \beta_3=0.999$) \\
\bottomrule
\end{tabular}
\label{tab:trainer2}
\end{table}

\subsection{Computational Resources}
\label{app:compute}

Both our method and the reproduced baseline are trained using
8 NVIDIA A100-SXM4-80GB GPUs on a server equipped with
Intel Xeon Platinum 8358 CPUs @ 2.60GHz.

For the four LIBERO tasks, we perform full-parameter fine-tuning for
30,000 training steps on video and 40,000 training steps on value/action,
which requires approximately 5 days.

For the real-world Piper manipulation data, we train one model per task
for 30,000 training steps on video and 30,000 training steps on
value/action, which requires approximately 3 days.







\end{document}